\documentclass[sigconf, preprint]{acmart}

\AtBeginDocument{%
  }

\setcopyright{none}
\copyrightyear{}
\acmYear{2026}
\acmDOI{}
\acmConference[]{}{}{}
\acmISBN{}

\usepackage{booktabs}

\usepackage{multirow}
\usepackage{amsmath}
\usepackage{graphicx}
\usepackage{microtype}

\begin{document}

\title{GUIDE: Interpretable GUI Agent Evaluation via Hierarchical Diagnosis}

\author{Yuwen Zhai$^{1*}$, Runze Li$^{2*}$, Liang Wang$^{1}$,  Nian Shi$^{1}$, Liwu Xu$^{1}$, \\
Wei Zhang$^{2,3}$, Ran Lin$^{1}$, Bo Xu$^{1}$, Benlei Cui$^{1}$\vspace{0.5em}}
\affiliation{\institution{$^{1}$Alibaba Group \quad $^{2}$School of Computer Science and Technology, East China Normal University \\
$^{3}$Shanghai Innovation Institute}
\country{}}
\thanks{$^{*}$Equal contribution.}

\renewcommand{\shortauthors}{Anonymous et al.}
\renewcommand{\authorsaddresses}{}

\begin{abstract}
Evaluating GUI agents presents a distinct challenge: trajectories are long, visually grounded, and open-ended, yet evaluation must be both accurate and interpretable.
Existing approaches typically apply a single holistic judgment over the entire action-observation sequence---a strategy that proves unreliable on long-horizon tasks and yields binary verdicts offering no insight into where or why an agent fails.
This opacity limits the utility of evaluation as a diagnostic tool for agent development.

We introduce GUIDE (GUI Understanding and Interpretable Diagnostic Evaluation), a framework that decomposes trajectory assessment into three sequential stages mirroring the compositional structure of GUI tasks.
\emph{Trajectory Segmentation} partitions the full trace into semantically coherent subtask units.
\emph{Subtask Diagnosis} evaluates each unit in context, assigning a completion verdict and generating a structured error analysis with corrective recommendations.
\emph{Overall Summary} aggregates per-subtask diagnoses into a task-level judgment.
By operating on bounded subtask segments rather than full trajectories, GUIDE mitigates the context overload that degrades existing evaluators as task complexity grows.

We validate GUIDE on three benchmarks: an industrial e-commerce dataset of 932 trajectories, AGENTREWARDBENCH spanning five web agent tasks with 1{,}302 trajectories, and AndroidBench for mobile device control.
Across all settings, GUIDE substantially outperforms existing evaluators---achieving up to 5.35 percentage points higher accuracy than the strongest baseline---while producing structured diagnostic reports that directly inform agent improvement.

\end{abstract}



\maketitle

\section{Introduction}

Driven by advances in large vision-language models (VLMs), GUI agents---AI systems that interact with software interfaces by perceiving screen states and executing actions---are increasingly deployed for complex, multi-step automation tasks across web browsers, mobile applications, and desktop environments~\cite{anthropic2024claude,openai2025operator,zheng2024seeact,he2024webvoyager}.
As these agents tackle longer and more consequential workflows, reliable evaluation of their performance becomes a foundational requirement for both research and deployment.

Evaluating GUI agent trajectories is inherently challenging.
Tasks are open-ended: the same goal can be accomplished through many different action sequences, rendering rule-based matching brittle and difficult to scale.
Trajectories are multimodal: a complete trace interleaves screenshots and natural-language actions over dozens of steps.
Ground-truth labels are expensive to obtain, requiring annotators to replay trajectories and exercise task-domain expertise.
These properties have motivated a shift toward LLM-as-judge approaches~\cite{zheng2023judging,pan2024autonomous,xue2025illusion,bhonsle2025autoeval}, which generalise across tasks without per-task rule engineering.

However, existing LLM-based evaluators share a design choice that limits both their reliability and utility: they evaluate the trajectory in a single monolithic call and return a single binary verdict.
This monolithic strategy exhibits two concrete failure modes.
First, as trajectory length grows, the evaluator faces increasing context overload: on our industrial benchmark, WebJudge's accuracy drops from 92.9\% on trajectories shorter than 10 steps to 75.0\% on trajectories of 50--80 steps---a degradation of nearly 18 percentage points.
Second, a binary success/failure verdict conveys no information about \emph{where} or \emph{why} the agent failed, which subtasks were completed correctly, or how performance could be improved.
An evaluation system with these limitations cannot serve as a meaningful diagnostic tool for agent development.

Hierarchical decomposition has been widely adopted on the \emph{execution} side of agent systems---for task planning, memory management, and self-improvement~\cite{shinn2023reflexion}---but the same principle has not been transferred to \emph{evaluation}.
Where some structure exists in existing evaluators, it either relies on manually pre-specified task milestones~\cite{ma2024agentboard} or coarser evaluation granularity without segmenting the trajectory itself.
GUIDE bridges this gap: rather than decomposing a task \emph{goal} to guide future execution, it decomposes a \emph{completed trajectory} to enable post-hoc diagnosis---evaluating each subtask segment independently against bounded context to achieve both robustness on long trajectories and structured, interpretable output.

\begin{figure*}[t]
  \centering
  \includegraphics[width=\linewidth]{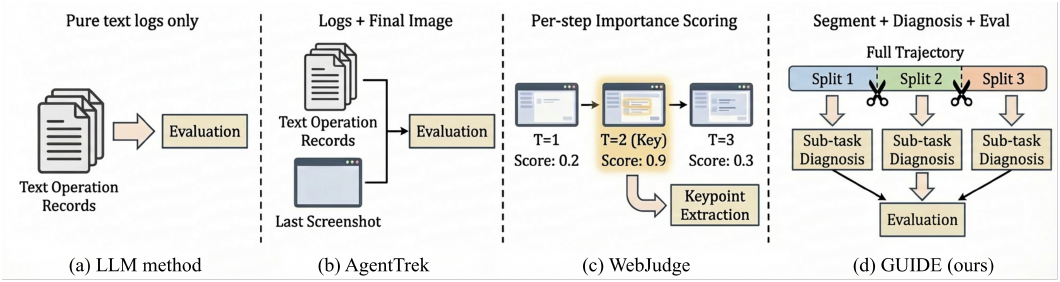}
  \caption{Comparison of GUI agent evaluation paradigms. From left to right: (1)~text-only LLM evaluators that ignore visual state; (2)~lightweight multimodal methods (AgentTrek) that append only the final screenshot; (3)~WebJudge, which retrieves key frames from the trajectory and evaluates them holistically in a single context; and (4)~GUIDE (ours), which decomposes the trajectory into subtask segments and applies structured diagnosis to each, combining full visual coverage with bounded per-call context.}
  \label{fig:method_comparison}
\end{figure*}

In this paper, we present \textbf{GUIDE} (\textbf{G}UI \textbf{U}nderstanding and \textbf{I}nterp-retable \textbf{D}iagnostic \textbf{E}valuation), a framework that addresses these limitations through structured trajectory decomposition.
GUIDE operates in three stages.
In \emph{Trajectory Segmentation}, the full trajectory is automatically partitioned into semantically coherent subtask units.
In \emph{Subtask Diagnosis}, each unit is evaluated independently, producing a completion verdict, a root-cause error analysis, and corrective recommendations.
In \emph{Overall Summary}, the per-subtask diagnoses are synthesised into a final task-level judgment.
This decompose-then-diagnose architecture reduces the effective context per evaluation step, improves robustness on long-horizon tasks, and produces a hierarchical, interpretable evaluation report.

We validate GUIDE against five baselines on three benchmarks.
On an industrial e-commerce dataset of 932 human-verified trajectories, GUIDE achieves 95.80\% accuracy, outperforming the strongest baseline by 5.35 percentage points.
On AGENTREWARDBENCH~\cite{lu2025agentrewardbench}, spanning five web agent benchmarks with 1{,}302 trajectories, GUIDE achieves 89.21\% precision. We further evaluate on the AndroidBench~\cite{pan2024autonomous}, where GUIDE robustly surpasses all prior methods, achieving a state-of-the-art accuracy of ${94.9\%_{\pm0.3}}$. Long-trajectory analysis further confirms that GUIDE's advantage grows with trajectory length, directly validating the decomposition design.

\noindent Our contributions are:
\begin{itemize}
  \item \textbf{GUIDE}, a three-stage evaluation framework that decomposes GUI agent trajectories into subtask units and generates structured, interpretable diagnoses comprising completion verdicts, root-cause error analyses, and corrective recommendations that are directly actionable for agent developers.
  \item A demonstration that decomposition-based evaluation is inherently robust to trajectory length: GUIDE maintains 93--98\% accuracy across all length groups, whereas the strongest baseline (WebJudge) degrades by 17.9 percentage points (pp) from short to long trajectories, and AgentTrek collapses by 47.3 pp.
  \item A structured failure analysis and diagnostic output that goes beyond binary success/failure verdicts by localizing, analyzing, and providing corrective guidance for issues at the step-level to answer \emph{where} the agent failed, \emph{why} it failed, and \emph{how to fix it}---capabilities absent from all existing GUI agent evaluators.  
  \item State-of-the-art results on three diverse benchmarks spanning industrial e-commerce (95.80\% accuracy), web agent meta-evaluation (89.21\% precision), and mobile device control (94.9\% accuracy).
\end{itemize}

\section{Related Work}
\subsection{GUI Agents}

Modern GUI agents use large vision-language models (VLMs) as their core reasoning engine to perceive screen states and execute actions across web, mobile, and desktop environments.
On the web, WebArena~\cite{zhou2024webarena} and VisualWebArena~\cite{koh2024visualwebarena} provide realistic benchmarks with complex multi-step tasks, while Mind2Web~\cite{deng2023mind2web} targets cross-website generalization.
For mobile platforms, Android in the Wild~\cite{rawles2023aitw} offers large-scale data for Android device control.
OSWorld~\cite{xie2024osworld} extends the landscape to desktop environments with real-world tasks spanning multiple operating systems.
SeeAct~\cite{zheng2024seeact} and WebVoyager~\cite{he2024webvoyager} demonstrate that VLMs can ground web actions directly in visual observations, and commercial systems such as Claude Computer Use~\cite{anthropic2024claude} and OpenAI Operator~\cite{openai2025operator} bring GUI automation to general users at scale.
As these agents are deployed on increasingly complex, long-horizon tasks spanning dozens of actions across multiple application states, the need for scalable, reliable, and \emph{interpretable} evaluation has grown correspondingly.

\subsection{GUI Agent Evaluation}
\textit{Rule-based evaluation.} Early evaluation relied on deterministic metrics such as URL matching, DOM state comparison, and database assertions~\cite{zhou2024webarena,xie2024osworld}.
While reproducible, these approaches require per-task rule engineering, are brittle to UI changes, and---as shown by AGENTREWARDBENCH~\cite{lu2025agentrewardbench}---tend to underestimate agent success rates.

\begin{figure*}[t]
  \centering
  \includegraphics[width=\linewidth]{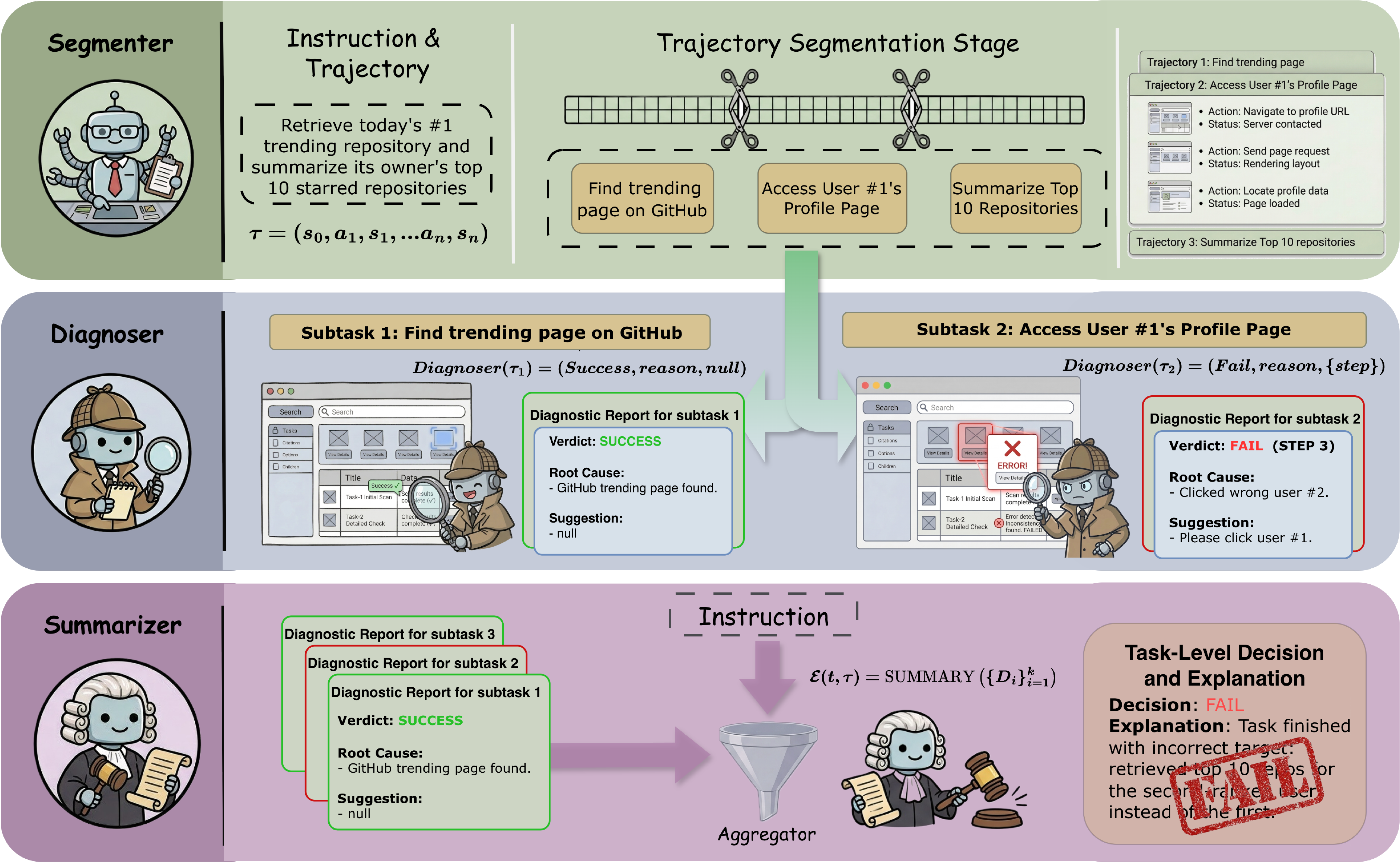}
  \caption{Overview of the GUIDE framework. Given a task description and a full agent trajectory,
  Module~1 (\emph{Trajectory Segmentation}) partitions the action-observation trace into
  semantically coherent subtask units. Module~2 (\emph{Subtask Diagnosis}) evaluates each unit
  independently, producing a structured diagnosis with a completion verdict, error analysis, and
  corrective recommendations. Module~3 (\emph{Overall Summary}) aggregates the per-subtask
  diagnoses into a final task-level judgment.}
  \Description{A pipeline diagram showing the GUIDE framework with three sequential modules: Module 1 segments a full GUI agent trajectory into subtask units, Module 2 diagnoses each subtask independently producing structured verdicts and error analyses, and Module 3 aggregates the diagnoses into a final task-level success or failure judgment.}
  \label{fig:framework}
\end{figure*}

\textit{LLM-as-judge.} The LLM-as-judge paradigm~\cite{zheng2023judging} offers task-agnostic evaluation through language model reasoning and has become the dominant alternative. Pan et al.~\cite{pan2024autonomous} pioneer autonomous evaluation for digital agents using VLMs without human annotation.
WebJudge~\cite{xue2025illusion} applies LLM-as-judge to key frames retrieved from web agent trajectories and reveals a ``progress illusion''---agents scoring $\sim$90\% on static benchmarks achieve only $\sim$30\% in live environments. Bhonsle et al.~\cite{bhonsle2025autoeval} propose Auto-Eval Judge, a general agentic framework for task completion assessment. Extensions to this paradigm include evaluator agents that actively inspect intermediate states~\cite{zhuge2025agentjudge} and graph-based analyses that surface critical decision points~\cite{qian2025webgrapheval}. In parallel, process reward models (PRMs) have begun to replace binary step correctness labels with continuous signals of future promise and progress~\cite{xi2025agentprm}.

\textit{Positioning of GUIDE.} All methods above share one or more of the following limitations: they evaluate the trajectory in a single context window (LLM-as-judge), require an execution environment or trained reward model (AgentPRM), or provide evaluation signals that are not directly interpretable by developers (graph statistics, scalar rewards).
GUIDE departs from these designs by decomposing each trajectory into subtask units before evaluation.
This decomposition reduces per-step context---the primary mechanism behind its robustness on long-horizon tasks---while producing structured, human-readable diagnostic reports that identify \emph{where} and \emph{why} an agent failed.
GUIDE requires no execution environment, no task-specific rules, and no model training, making it immediately deployable as a zero-shot evaluator.

Figure~\ref{fig:method_comparison} situates GUIDE within the landscape of existing evaluators along the axis of input modality and evaluation granularity.
Purely text-based LLM methods rely on action logs alone and cannot reason about visual state.
Methods such as AgentTrek and Autonomous Evaluation augment the text trajectory with a single final screenshot, providing minimal visual grounding at the cost of losing all intermediate visual context. WebJudge retrieves key frames from the trajectory and evaluates them holistically in a single context window, which becomes unreliable as trajectory length grows.
GUIDE is the only approach that combines full multimodal input with fine-grained subtask decomposition, enabling both visual grounding at each step and bounded context per evaluation call.

\begin{figure*}[t]
  \centering
  \includegraphics[width=\textwidth]{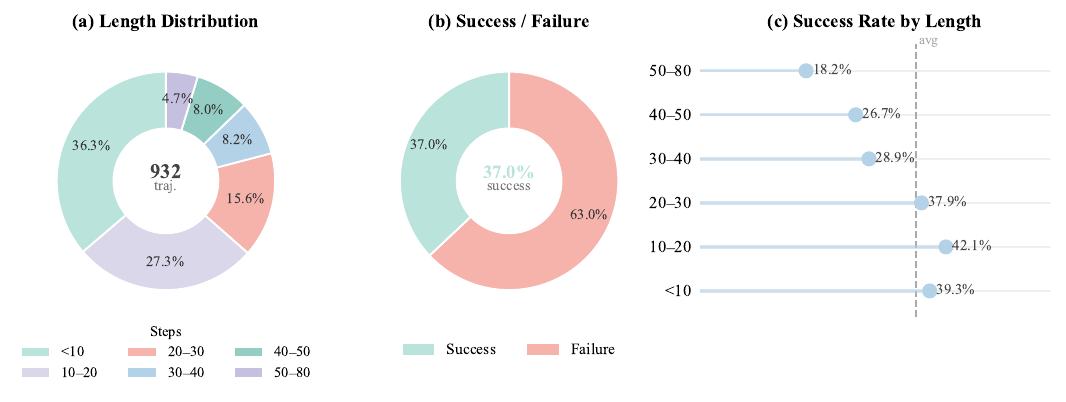}
  \caption{Statistics of the industrial e-commerce dataset. \textbf{(a)}~Trajectory length distribution across six groups. \textbf{(b)}~Overall success/failure ratio (37.0\% vs.\ 63.0\%). \textbf{(c)}~Per-group success rate; success decreases monotonically from 39.4\% ($<$10 steps) to 18.2\% (50--80 steps), confirming that longer trajectories are inherently more challenging.}
  \label{fig:dataset}
\end{figure*}

\subsection{Hierarchical Decomposition in LLM Agent Systems}
Hierarchical decomposition is the strategy of breaking complex tasks into manageable subunits.

\textit{Task Planning.} Hierarchical approaches decompose high-level goals into executable subgoals. SayCan~\cite{ahn2022saycan} combines LLM-based semantic planning with affordance models to ground abstract goals in physical feasibility. DEPS~\cite{wang2023deps} and Voyager~\cite{wang2023voyager} demonstrate hierarchical planning in open-world environments. StructuredAgent~\cite{lobo2025structuredagent} extends this to web navigation using AND/OR trees to represent task structure.

\textit{Reasoning Enhancement}, hierarchical structures organize the problem-solving process itself. Tree of Thoughts~\cite{yao2023tree} enables LLMs to explore multiple reasoning branches through tree-structured search, while Plan-and-Solve Prompting~\cite{wang2023planandsolveprompting} proposes a two-stage approach that first generates a plan by decomposing the problem into subgoals, then executes each step sequentially.
These methods improve reasoning accuracy by introducing structured exploration over flat chain-of-thought approaches.

\textit{Memory Management.} Hierarchical organization addresses the challenge of long-term information retention. Generative Agents~\cite{park2023generative} introduces a three-tier memory architecture---observation stream, reflections, and plans---enabling agents to extract high-level insights from raw experiences. HiAgent~\cite{hu2025hiagent} proposes hierarchical working memory management that compresses completed subtasks into summaries, mitigating context overload during long-horizon execution.

Recent work has also explored hierarchical approaches in error detection and recovery. Reflexion~\cite{shinn2023reflexion} uses verbal feedback to guide iterative refinement, while BacktrackAgent~\cite{wu2025backtrackagent} employs a three-component architecture (Verifier, Judger, Reflector) for detecting and recovering from GUI action errors during execution. GUI-Critic-R1~\cite{wanyan2025guicritic} shifts this paradigm further by introducing pre-operative diagnosis---detecting potential errors before action execution.

\section{The GUIDE Framework}
\subsection{Problem Formulation}

A GUI agent trajectory is a sequence of interleaved observations and actions:
\begin{equation}
  \tau = (s_0,\, a_1, s_1,\, a_2, s_2,\, \ldots,\, a_n, s_n),
\end{equation}
where $s_i$ denotes the agent's screen observation at step $i$ (typically a screenshot, optionally augmented with an accessibility tree), and $a_i$ denotes the action executed at step $i$ (e.g., click, type, scroll).
Given a task description $t$ and trajectory $\tau$ of $n$ steps, the goal of an evaluation function $\mathcal{E}(t, \tau) \in \{0, 1\}$ is to determine whether the agent successfully completed $t$.

Most existing evaluators implement $\mathcal{E}$ as a single model call over the full input $(t, \tau)$, concatenating all observations and actions into one context window.
GUIDE instead decomposes evaluation into three sequential stages:
\begin{equation}
  \mathcal{E}(t, \tau) = \textsc{Summary}\!\left(\{D_i\}_{i=1}^{k}\right),
  \quad D_i = \textsc{Diagnose}(t_i,\, \tau_i),
  \label{eq:guide}
\end{equation}
where $\{(t_i, \tau_i)\}_{i=1}^{k}$ is a partition of $\tau$ into $k$ subtask segments produced by the Trajectory Segmentation Module (Module~1), and $D_i$ is the structured diagnosis of the $i$-th subtask from the Subtask Diagnosis Module (Module~2). $\mathcal{E}(t, \tau)$ represents the overall assessment from the Overall Summary Module (Module~3). Figure~\ref{fig:framework} illustrates the full pipeline.

\subsection{Module 1: Trajectory Segmentation}

\paragraph{Goal.}
Given the task description $t$ and the action sequence $\tau$, Module~1 identifies a set of segment boundaries $\mathcal{B} = \{0 = b_0 < b_1 < \cdots < b_k = n\}$ that partition $\tau$ into $k$ non-overlapping subtask segments:
\begin{equation}
  \tau_i = (s_{b_{i-1}},\, a_{b_{i-1}+1}, s_{b_{i-1}+1},\, \ldots,\, a_{b_i}, s_{b_i}).
\end{equation}
Each segment $\tau_i$ is accompanied by an inferred subtask description $t_i$ that characterises the goal of that segment within the broader task.

\paragraph{Design.}
Segmentation is performed by a language model that receives $t$ and the textual representation of the full action sequence (action summary, without screenshots), and produces the boundaries $\mathcal{B}$ together with the per-segment descriptions $\{t_i\}$ in a structured output format. We deliberately exclude screenshots from Module~1: the action sequence alone suffices to capture the semantic transitions between subtasks, and omitting images keeps the segmentation step lightweight.

\paragraph{Why segmentation matters.}
By decomposing a trajectory of $n$ steps into $k$ segments, the maximum context any single subsequent module must process shrinks from $O(n)$ to $O(n/k)$.
For a 60-step trajectory segmented into six units of ten steps each, the per-evaluation context load is reduced by an order of magnitude.
This reduction is the primary mechanism behind GUIDE's robustness on long-horizon tasks.
Moreover, segmentation introduces a natural abstraction layer: Module~2 receives subtask descriptions alongside the raw steps, enabling it to evaluate each segment against a local objective rather than requiring it to maintain a mental model of the entire task history.
This locality property also makes GUIDE's evaluations more interpretable, as each diagnosis is anchored to a specific subtask with a clear scope.
We provide quantitative evidence for segmentation quality in Section~\ref{sec:seg_quality}, where an MLLM-based evaluator confirms that 99.4\% of produced subtasks are usable for downstream diagnosis.


\subsection{Module 2: Subtask Diagnosis}

\paragraph{Goal.}
For each subtask segment $(t_i, \tau_i)$, Module~2 produces a structured diagnosis:
\begin{equation}
  D_i = (v_i,\; e_i,\; c_i),
\end{equation}
where $v_i \in \{\textit{success},\, \textit{partial},\, \textit{fail}\}$ is the completion verdict for the subtask, $e_i$ is a natural-language error analysis identifying the root cause of any failure or deficiency, and $c_i$ is a set of step-level corrective recommendations.
The three-class verdict distinguishes between subtasks that are fully completed, partially completed with residual errors, and those that fail outright---a distinction that allows Module~3 to reason more precisely about overall task success.

\paragraph{Design.}
Module~2 is multimodal: it receives the subtask description $t_i$, the full list of subtask descriptions $\{t_j\}_{j=1}^k$ for context, the action sequence for the segment, and the corresponding screenshots $\{s_{b_{i-1}}, \ldots, s_{b_i}\}$ along with a final-state screenshot.
A vision-language model processes this input and generates $(v_i, e_i, c_i)$ via a structured JSON output schema. The schema enforces a chain-of-thought reasoning trace before the model commits to a verdict, and requires the enumeration of any problematic steps, each accompanied by an analysis of its root cause and a suggested fix---preventing the model from issuing a verdict without supporting analysis. Providing the full subtask list as context ensures the model does not re-evaluate steps already addressed in prior subtasks, avoiding double-counting of completed objectives.

\paragraph{Interpretability.}
The diagnostic triple $(v_i, e_i, c_i)$ is the central contribution of Module~2 and the primary source of GUIDE's interpretability.
Unlike a bare binary verdict, it answers three questions that matter for agent development: \emph{did this subtask succeed?}, \emph{why did it fail?}, and \emph{what should the agent have done instead?}
The structured output can be surfaced directly to developers for error analysis, or injected back into the agent as corrective context for refinement.

\begin{table}[t]
  \centering
  \caption{Results on the industrial e-commerce dataset. Best per column in \textbf{bold}; $\uparrow$ = higher is better.}
  \label{tab:ecommerce}
  \setlength{\tabcolsep}{4.5pt}
  \begin{tabular}{lrrrrr}
    \toprule
    Method & Acc$\uparrow$ & P$\uparrow$ & R$\uparrow$ & F1$\uparrow$ \\
    \midrule
    AgentTrek          & 65.02 & 51.42 & \textbf{100.00} & 67.91 \\
    Plan\&Solve-like       & 82.94 & 72.91 & 85.80 & 78.83 \\
    Autonomous Eval    & 84.44 & 77.93 & 80.87 & 79.37 \\
    WebJudge           & 90.45 & 84.04 & 91.59 & 87.66 \\
    \midrule
    \textbf{GUIDE (Ours)} & \textbf{95.80} & \textbf{95.00} & 93.62 & \textbf{94.31} \\
    \bottomrule
  \end{tabular}
\end{table}

\subsection{Module 3: Overall Summary}

\paragraph{Goal.}
Module~3 aggregates the $k$ subtask diagnoses $\{D_1, \ldots, D_k\}$ into a final task-level verdict $\mathcal{E}(t, \tau) \in \{0, 1\}$.

\paragraph{Design.}
Rather than applying a fixed aggregation rule (e.g., succeed only if all subtasks are correct), Module~3 uses a language model that reasons holistically over the diagnoses.
This is important because real-world GUI tasks do not always decompose cleanly: an agent may fail an intermediate subtask but recover, or may complete all subtasks nominally yet miss the overarching goal due to accumulated small errors.
The model receives the task description $t$, the per-subtask diagnostic reasoning traces, and the step-level issue lists for each subtask, then produces the final verdict with a brief justification.
The subtask summaries generated by Module~1 are provided as secondary reference only; the model is instructed to prioritise the diagnostic evidence and the original task instruction when they conflict.

\paragraph{Design rationale.}
Several design choices merit explicit justification. First, Module~1 uses text-only input rather than screenshots because action sequences alone are sufficient for semantic boundary detection. Moreover, including the full image sequence for long trajectories is not only computationally expensive but also highly prone to API failures due to excessive context length. This lightweight design proves both effective and high-quality, as validated by the segmentation quality analysis in Section~\ref{sec:seg_quality}. Second, Module 2's primary role is to identify specific problematic steps within each subtask. This step-level diagnosis is crucial, as it directly informs Module 3's ability to distinguish between minor, recoverable issues and critical, task-level failures when issuing the final verdict.

\section{Experiments}

\subsection{Experimental Setup}
\begin{table*}[t]
  \centering
  \caption{Results on AGENTREWARDBENCH (1,302 trajectories, 5 web benchmarks).
  Methods marked $\dagger$ are taken from~\cite{lu2025agentrewardbench}.
  Per-benchmark columns report Precision (\%).}
  \label{tab:agentrewardbench}
  \setlength{\tabcolsep}{4pt}
  \begin{tabular}{lr rrrrr}
    \toprule
    & \multicolumn{1}{c}{Overall} & \multicolumn{5}{c}{Per-Benchmark Precision$\uparrow$} \\
    \cmidrule(lr){2-2} \cmidrule(lr){3-7}
    Method & P$\uparrow$ & Asst. & VWA & WA & W.Arena & W.Arena\textsuperscript{++} \\
    \midrule
    Rule-based$^\dagger$         & 83.8  & 25.0 & \textbf{85.2} & 79.0 & 100.0 & 83.3 \\
    NNetNav$^\dagger$            & 52.5  & 20.8 & 54.5 & 54.3 & 77.3 & 43.2 \\
    Autonomous Eval$^\dagger$    & 67.6  & 83.3 & 61.2 & 67.6 & 96.4 & 59.3 \\
    GPT-4o (A11y Tree)$^\dagger$ & 69.8  & 77.8 & 63.0 & 70.2 & 94.6 & 63.0 \\
    Claude 3.7 Sonnet$^\dagger$  & 69.4  & 71.4 & 64.8 & 69.3 & 85.3 & 66.7 \\
    WebJudge (o4-mini)$^\dagger$ & 82.0  & \textbf{100.0} & 74.5 & 81.2 & 100.0 & 90.0 \\
    \midrule
    \textbf{GUIDE} (o4-mini)     & 86.67 & \textbf{100.0} & 76.6 & 89.7 & \textbf{100.0} & 87.0 \\
    \textbf{GUIDE}     & \textbf{89.21} & \textbf{100.0} & 78.9 & \textbf{91.7} & \textbf{100.0} & \textbf{90.9} \\
    \bottomrule
  \end{tabular}
\end{table*}
\paragraph{Datasets.}
We evaluate on three benchmarks.

\textbf{Industrial E-commerce Dataset.}
We collect 932 trajectories from an industrial e-commerce platform, covering multi-step workflows including order management, product search, and product listing.
Each trajectory is labelled by human annotators as success or failure; the dataset contains 345 successes (37.0\%) and 587 failures (63.0\%), spanning 633 distinct task instructions---reflecting substantial diversity in user intent.
Trajectory length ranges from 5 to 80 steps, with a mean of approximately 19 steps. Figure~\ref{fig:dataset} summarises the dataset distribution by trajectory length. It spans a wide range of trajectory lengths (a), and each length bracket contains a healthy mix of successful and failed outcomes (c). This distribution ensures a challenging yet fair benchmark, free from biases at any specific complexity level.


\textbf{AGENTREWARDBENCH}~\cite{lu2025agentrewardbench} is a meta-evaluation benchmark comprising 1,302 web agent trajectories drawn from five established benchmarks: AssistantBench, VisualWebArena (VWA), WebArena (WA), WorkArena, and WorkArena\textsuperscript{++}.
Each trajectory carries a ground-truth label verified by the respective benchmark authors.

\textbf{AndroidBench.}
We evaluate on the dataset released by Pan et al.~\cite{pan2024autonomous}, which consists of 120 tasks randomly sampled from the AitW test set~\cite{rawles2023aitw}. The evaluation uses an Android emulator, with human judgments of trajectory success serving as the oracle. We compare GUIDE against the evaluators reported in~\cite{pan2024autonomous} on this established benchmark.
\begin{table}[t]
  \centering
  \caption{Accuracy (\%) on AndroidBench~\cite{pan2024autonomous} (480 trajectories, 4 agent policies).
  Methods marked $\dagger$ are from Pan et al.~\cite{pan2024autonomous} and use different backbones.
  Remaining methods use \texttt{gemini-3.0-flash}; all values show mean$_{\pm\text{std}}$ over 3 independent runs.}
  \label{tab:mobile}
  \setlength{\tabcolsep}{2.5pt}
  \begin{tabular}{lrrrrr}
    \toprule
    Method & Overall & AitW & AutoUI-b & AutoUI-l & CogAgent \\
    \midrule
    QWen-VL$^\dagger$        & 70.2 & — & — & — & — \\
    GPT-4V$^\dagger$         & 90.6 & — & — & — & — \\
    Cap.+GPT-4$^\dagger$     & 89.8 & — & — & — & — \\
    Cap.+Mixtral$^\dagger$   & 92.9 & — & — & — & — \\
    \midrule
    AgentTrek             & $92.4_{\pm0.1}$ & $86.1_{\pm0.5}$ & $93.9_{\pm0.5}$ & $99.2_{\pm0.0}$ & $90.3_{\pm0.5}$ \\
    WebJudge              & $92.6_{\pm0.5}$ & $83.6_{\pm1.0}$ & $93.3_{\pm1.7}$ & $\mathbf{100.0_{\pm0.0}}$ & $93.3_{\pm0.8}$ \\
    \textbf{GUIDE (Ours)} & $\mathbf{94.9_{\pm0.3}}$ & $\mathbf{85.6_{\pm1.9}}$ & $\mathbf{98.1_{\pm0.5}}$ & $\mathbf{100.0_{\pm0.0}}$ & $\mathbf{95.8_{\pm0.9}}$ \\
    \bottomrule
  \end{tabular}
\end{table}
\paragraph{Baselines.}
We compare against the following evaluators.
\textbf{Autonomous Evaluation}~\cite{pan2024autonomous} assesses task completion using the final screenshot together with the preceding action descriptions, without human annotation.
\textbf{WebJudge}~\cite{xue2025illusion} applies LLM-as-judge to key frames retrieved from the web agent trajectory.
\textbf{AgentTrek} evaluates task completion using the final screenshot together with the full action sequence as context.
\textbf{Rule-based} uses deterministic matching against annotated expected outcomes.
On AGENTREWARDBENCH, GUIDE is evaluated with two backbone models (\texttt{o4-mini} and \texttt{gemini-3.0-flash}); results marked with $\dagger$ are taken directly from~\cite{lu2025agentrewardbench}.
\paragraph{Metrics.}
We report Accuracy (Acc), Precision (P), Recall (R), and F1 score (F1).
All metrics treat task success as the positive class.

\paragraph{Implementation.}
All three GUIDE modules use \texttt{gemini-3.0-flash} as the backbone.
All modules are prompted to produce JSON output; responses are parsed with a robust extraction procedure that handles code blocks and malformed output, with up to 10 retries with exponential backoff on parse failure.
All modules use the default API temperature.

\subsection{Main Results}

\paragraph{Industrial E-commerce Dataset.}
Table~\ref{tab:ecommerce} presents the main comparison.
GUIDE achieves 95.80\% accuracy and an F1 of 94.31\%, outperforming the strongest baseline (WebJudge) by 5.35 percentage points in accuracy. Autonomous Evaluation reaches 84.44\% accuracy, reflecting the difficulty of holistic trajectory assessment on this dataset. The Plan\&Solve-like baseline, which employs a top-down strategy of first planning subtasks and then aligning them to the trajectory, achieves a lower accuracy of 82.94\%. This suggests its rigid, pre-defined plan struggles to handle the non-linear and often unpredictable nature of real-world agent executions. AgentTrek achieves perfect recall (100\%) at the expense of precision (51.42\%) by classifying most trajectories as successes, indicating a near-degenerate strategy and resulting in 65.02\% accuracy.
\begin{table*}[t]
  \centering
  \caption{Accuracy (\%) by trajectory length on the industrial e-commerce dataset.
  GUIDE maintains stable performance across all groups while baselines degrade as trajectories grow longer.}
  \label{tab:longtraj}
  \setlength{\tabcolsep}{6pt}
  \begin{tabular}{lrrrrrr}
    \toprule
    Method & $<$10 steps & 10--20 & 20--30 & 30--40 & 40--50 & 50--80 \\
    \midrule
    AgentTrek       & 81.4 & 62.2 & 55.9 & 54.0 & 48.0 & 34.1 \\
    Plan\&Solve-like   & 94.1 & 92.1 & 88.3 & 67.1 & 40.0 & 27.3 \\
    Autonomous Eval & 87.9 & 85.0 & 75.2 & 85.5 & 85.3 & 81.8 \\
    WebJudge        & 92.9 & 92.5 & 89.0 & 89.5 & 85.3 & 75.0 \\
    \midrule
    \textbf{GUIDE (Ours)} & \textbf{97.6} & \textbf{94.9} & \textbf{93.8} & \textbf{96.1} & \textbf{95.9} & \textbf{93.2} \\
    \bottomrule
  \end{tabular}
\end{table*}
\paragraph{AGENTREWARDBENCH}
Table~\ref{tab:agentrewardbench} presents results on the web agent meta-evaluation benchmark.
We evaluate GUIDE with two backbone models: \texttt{o4-mini} and \texttt{gemini-3.0-flash} (default).
GUIDE achieves the highest overall precision (89.21\%) among all evaluated methods, surpassing the strongest reported baseline, WebJudge (o4-mini) (82.0\%), by 7.2 percentage points.
GUIDE (o4-mini) also outperforms WebJudge (o4-mini) in precision (86.67\% vs.\ 82.0\%).
Both GUIDE variants achieve 100\% precision on AssistantBench and WorkArena, and maintain high precision across VisualWebArena (76.6--78.9\%), WebArena (89.7--91.7\%), and WorkArena\textsuperscript{++} (87.0--90.9\%), demonstrating consistent reliability across diverse task domains.
The best prior system reported in~\cite{lu2025agentrewardbench}---GPT-4o with accessibility tree---achieves 69.8\% precision; GUIDE surpasses it by 19.4 percentage points.
Notably, no single prior evaluator dominates all five sub-benchmarks simultaneously~\cite{lu2025agentrewardbench}, whereas both GUIDE variants maintain competitive or leading precision on every sub-benchmark, suggesting that the decompose-then-diagnose strategy yields more balanced generalisation across task domains.

\paragraph{AndroidBench}
Table~\ref{tab:mobile} reports results on AndroidBench~\cite{pan2024autonomous}, evaluated over 480 trajectories (120 tasks $\times$ 4 agent policies: AitW human demos, AutoUI-base, AutoUI-large, and CogAgent).
To address concerns about result stability on this benchmark, we run each LLM-based evaluator three times independently and report mean accuracy with standard deviation (all powered by \texttt{gemini-3.0-flash} for fairness).
GUIDE achieves $94.9 \pm 0.3$\% mean accuracy, outperforming both same-backbone baselines (WebJudge $92.6 \pm 0.5$\%, AgentTrek $92.4 \pm 0.1$\%) and the best Autonomous Evaluation variant from Pan et al.~\cite{pan2024autonomous} (Captioner + Mixtral, 92.9\%).
The narrow standard deviation of GUIDE ($\pm 0.3$) compared to WebJudge ($\pm 0.5$) further indicates that GUIDE's advantage is consistent rather than a statistical artefact.
The per-policy breakdown (averaged over three runs) shows that GUIDE achieves perfect accuracy on AutoUI-large and the highest accuracy on AutoUI-base (98.1\%) and CogAgent (95.8\%), demonstrating robust evaluation across agent architectures of varying capability.
Interestingly, the AitW human demo subset proves most challenging for all evaluators, including GUIDE (85.6\%).
Human demonstrations exhibit more naturalistic and exploratory behaviour---including hesitations, corrections, and non-optimal paths---that can confuse evaluators trained on or prompted for more systematic agent trajectories.
Despite this, GUIDE's overall accuracy remains the highest among all methods, confirming that the decomposition strategy generalises across both automated and human-generated trajectories.


\subsection{Long-Trajectory Analysis}

A central claim of GUIDE is that trajectory decomposition improves robustness as task complexity grows.
Table~\ref{tab:longtraj} tests this claim directly by partitioning the industrial e-commerce dataset into six trajectory-length groups and reporting accuracy for each method.

The results strongly support the decomposition hypothesis.
As trajectory length increases, existing evaluators degrade substantially: WebJudge's accuracy falls from 92.9\% on short trajectories ($<$10 steps) to 75.0\% on long ones (50--80 steps)---a drop of 17.9 percentage points. AgentTrek collapses even more severely, from 81.4\% to 34.1\% (-47.3 pp). Autonomous Evaluation exhibits more moderate degradation but still drops from 87.9\% to 81.8\% on the longer groups. The Plan\&Solve-like baseline, which employs a top-down planning strategy, suffers an even more dramatic collapse. Its accuracy plummets from a high of 94.1\% on short tasks to a mere 27.3\% on the longest ones. This failure highlights the brittleness of its rigid, pre-defined plan, which struggles to align with the non-linear and often unpredictable nature of real-world agent executions. GUIDE, by contrast, maintains stable accuracy across all length groups, ranging from 93.2\% to 97.6\%.
The advantage over WebJudge, the strongest baseline, grows monotonically with trajectory length: from 4.7 pp on short trajectories to 18.2 pp on the 50--80 step group.
This widening gap directly validates GUIDE's core hypothesis: by decomposing long trajectories into bounded subtask segments, the effective context per evaluation step remains constant regardless of total trajectory length, whereas monolithic evaluators must fit an ever-growing context into a single inference call.
The stability of GUIDE across all length groups also suggests that the segmentation module produces consistent subtask boundaries even for complex, multi-phase workflows.


\subsection{Ablation Study}
To quantify the contribution of each module, we ablate GUIDE on the industrial e-commerce dataset under four configurations:
(a)~\textbf{w/o Seg}: the full trajectory is passed directly to Module~2 without segmentation;
(b)~\textbf{w/o Diag}: trajectories are segmented but each subtask receives only a binary verdict with no error analysis or corrective output;
(c)~\textbf{w/o Sum}: the final verdict is determined by a hard rule (failure if any subtask is non-success) without Module~3;
(d)~\textbf{Naive}: a single LLM call evaluates the full trajectory with no decomposition and no structured output.
\begin{table}[h]
  \centering
  \caption{Ablation study on the industrial e-commerce dataset.}
  \label{tab:ablation}
  \begin{tabular}{lrrrr}
    \toprule
    Variant & Acc$\uparrow$ & P$\uparrow$ & R$\uparrow$ & F1$\uparrow$ \\
    \midrule
    Naive (single LLM call)  & 85.61 & 78.59 & 84.06 & 81.23 \\
    w/o Seg                  & 74.14 & 61.40 & 81.16 & 69.91 \\
    w/o Diag                 & 93.22 & 88.32 & 94.20 & 91.16 \\
    w/o Sum                  & 87.41 & \textbf{95.60} & 69.28 & 80.34 \\
    \midrule
    \textbf{GUIDE (full)}    & \textbf{95.80} & 95.00 & \textbf{93.62} & \textbf{94.31} \\
    \bottomrule
  \end{tabular}
\end{table}

Table~\ref{tab:ablation} reports the results.
Removing segmentation (w/o Seg) causes the largest single drop: accuracy falls to 74.14\%, 21.66 percentage points below the full model and even below the Naive baseline.
This counter-intuitive result arises because Module~2's structured diagnostic prompt is optimised for subtask-length inputs; when applied to a full trajectory without prior segmentation, the prompt's diagnostic structure becomes a liability rather than an asset, confusing the model more than a simple holistic prompt would.
Removing error analysis (w/o Diag) yields a smaller but consistent drop to 93.22\%, confirming that the structured diagnosis contributes beyond the verdict alone.
Removing the summary module (w/o Sum) and replacing it with a hard all-or-nothing rule reduces accuracy to 87.41\%, primarily through a severe drop in recall (69.28\%): the rigid rule fails to handle trajectories where an agent recovers from an intermediate subtask failure and ultimately completes the task.
These results validate each module's contribution and confirm that the decompose-then-diagnose design is essential rather than incidental.
\subsection{Segmentation Quality Validation}
\label{sec:seg_quality}
A potential concern with GUIDE is whether Module~1 consistently produces high-quality subtask segments: poorly-formed segments could degrade downstream diagnosis regardless of the overall pipeline design.
We address this by constructing an MLLM-based segmentation quality evaluator that independently assesses each produced subtask segment on two dimensions: (1)~\emph{coherence \& boundary quality}---whether the steps form a semantically self-contained unit with natural boundaries---and (2)~\emph{description--behaviour alignment}---whether the inferred subtask description accurately reflects the observed actions.
The evaluator produces an integer score from 1 (unusable: degenerate split that would mislead downstream diagnosis) to 5 (highly usable: coherent segment with accurate description).
The full evaluation prompt is provided in Appendix.

\paragraph{Human validation.}
To verify that the MLLM evaluator is reliable, we sampled 200 subtasks and asked a human annotator to independently label each as \emph{usable} (score $\ge 4$) or \emph{problematic} (score $\le 3$).
Inter-rater agreement between the human binary labels and the binarised MLLM scores reached Cohen's $\kappa = 0.89$, indicating strong agreement and confirming that the automated evaluator serves as a credible proxy for human judgment.

\paragraph{Results.}
We apply the evaluator to all 3.3k subtasks produced from all industrial e-commerce trajectories.
Table~\ref{tab:seg_quality} reports the score distribution.
The vast majority of subtasks receive scores of 4 or 5 (99.4\%), indicating that Module~1 consistently produces coherent, well-described segments.
Only 0.6\% subtasks fall below 4, confirming that degenerate splits are rare and are unlikely to materially affect downstream evaluation quality.
These results provide quantitative evidence that GUIDE's segmentation step is robust and reliably produces inputs suitable for diagnosis by Module~2.

\begin{table}[t]
  \centering
  \caption{Segmentation quality score distribution across 3.3k subtasks on the industrial e-commerce dataset, as assessed by the MLLM evaluator.}
  \label{tab:seg_quality}
  \setlength{\tabcolsep}{5pt}
  \begin{tabular}{clrr}
    \toprule
    Score & Quality Level & \% \\
    \midrule
    5 & Highly Usable  & 96.9 \\
    4 & Usable          &  2.5 \\
    3 & Minor Issues    &  0.3 \\
    2 & Risky           &  0.2 \\
    1 & Unusable        &  0.1 \\
    \midrule
    \multicolumn{2}{l}{\textbf{Usable ($\ge 4$)}} & \textbf{99.4} \\
    \bottomrule
  \end{tabular}
\end{table}

\subsection{Discussion}
\paragraph{When does GUIDE fail?}
Among the 39 misclassified trajectories on the e-commerce dataset (4.2\% error rate), the dominant source of error is \emph{domain-specific business knowledge}.
GUIDE is a training-free, zero-shot framework: it is not fine-tuned on any domain-specific data and relies entirely on the general reasoning capabilities of the backbone VLM.
As a result, evaluation on tasks that require knowledge of platform-specific conventions---such as the implicit completion criteria for an order-management workflow or the expected final state of a supplier-facing listing operation---can be misjudged when the task description alone does not fully encode these domain norms.
This is an inherent trade-off of training-free evaluation: GUIDE achieves strong generalisation across domains without requiring labelled data or task-specific rules, at the cost of occasional errors on cases where domain expertise is indispensable.
Such errors could be substantially reduced by providing GUIDE with a concise domain-knowledge preamble in the prompt---a lightweight adaptation that requires no model retraining.

\section{Conclusion}

We presented GUIDE, an evaluation framework for GUI agent trajectories that replaces monolithic assessment with a three-stage decomposition pipeline.
Trajectory Segmentation partitions a full trace into semantically coherent subtask units; Subtask Diagnosis evaluates each unit independently, generating a structured report comprising a completion verdict, a root-cause error analysis, and corrective recommendations; and Overall Summary synthesises the per-subtask diagnoses into a final task-level judgment.
By operating on bounded subtask segments rather than full trajectories, GUIDE reduces per-evaluation context and improves robustness as task complexity grows.

Experiments across three benchmarks demonstrate consistent improvements over existing evaluators.
On an industrial e-commerce dataset, GUIDE achieves 95.80\% accuracy---5.35 percentage points above the strongest baseline---while maintaining stable performance across trajectory lengths ranging from fewer than 10 to over 50 steps.
On AGENTREWARDBENCH and AndroidBench, GUIDE attains the highest accuracy among all compared methods. Beyond accuracy, GUIDE produces hierarchical diagnostic output that identifies not only \emph{whether} an agent failed, but \emph{where} and \emph{why}, enabling targeted remediation.



\bibliographystyle{ACM-Reference-Format}
\bibliography{main}


\end{document}